\documentclass[letterpaper, 10 pt, conference]{ieeeconf}  

\IEEEoverridecommandlockouts                              

\usepackage{xcolor}
\usepackage{float}
\usepackage{amsmath}
\usepackage{amssymb}
\usepackage{multirow}
\usepackage{booktabs}
\usepackage{pifont}
\usepackage{tablefootnote}
\usepackage[abbreviations]{glossaries-extra}
\usepackage{tikz}
\usepackage{pgfplots}
\pgfplotsset{compat=newest}
\usepackage{caption}
\usepackage{tabularx}
\usepackage{subcaption}
\usepackage{hyperref}

\usepackage{graphicx}

\newcommand{\cmark}{\ding{51}}
\newcommand{\xmark}{\ding{55}}

\definecolor{mumblue}{RGB}{0,100,222}  
\definecolor{mumred}{RGB}{220,33,77} 
\definecolor{mumgreen}{RGB}{0,140,0} 
\definecolor{mumpurple}{RGB}{102,0,102} 
\definecolor{mumorange}{RGB}{255,102,0} 
\definecolor{mumteal}{RGB}{55,200,171} 
\definecolor{darkgrey}{RGB}{70,70,70} 
\definecolor{grey}{RGB}{150,150,150} 
\definecolor{lightgrey}{RGB}{225,225,225} 
\definecolor{lightgray}{RGB}{225,225,225} 
\definecolor{mumteallogo}{RGB}{45,198,214} 
\definecolor{mumbluelogo}{RGB}{24,59,101} 

\definecolor{finn}{RGB}{255, 0, 0}
\definecolor{timon}{RGB}{0, 0, 255} 
\definecolor{quantao}{RGB}{255, 0, 255}
\definecolor{olov}{RGB}{0, 155, 155}

\title{\LARGE \bf
One Map to Find Them All: Real-time Open-Vocabulary Mapping\\for Zero-shot Multi-Object Navigation
}

\author{Finn Lukas Busch, Timon Homberger, Jesús Ortega-Peimbert, Quantao Yang, and Olov Andersson
\thanks{*This work was partially supported by the Wallenberg AI, Autonomous Systems and Software Program (WASP) funded by the Knut and Alice Wallenberg Foundation.}
\thanks{The authors are with the Division of Robotics, Perception, and Learning, KTH Royal Institute of Technology, Sweden. Contact: \texttt{\{flbusch, timonh, jgop, quantao, olovand\}@kth.se}{\tt\small}}%
}
\renewcommand{\baselinestretch}{0.991}
\graphicspath{{figures/}} 

\begin{document}

\maketitle
\thispagestyle{empty}
\pagestyle{empty}

\begin{abstract}
%
The capability to efficiently search for  objects in complex environments is fundamental for many real-world robot applications. Recent advances in open-vocabulary vision models have resulted in semantically-informed object navigation methods that allow a robot to search for an arbitrary object without prior training. However, these zero-shot methods have so far treated the environment as unknown for each consecutive query. In this paper we introduce a new benchmark for zero-shot multi-object navigation,  allowing the robot to leverage information gathered from previous searches to more efficiently find new objects. To address this problem we build a reusable open-vocabulary feature map tailored for real-time object search. We further propose a probabilistic-semantic map update that mitigates common sources of errors in semantic feature extraction and leverage this semantic uncertainty for informed multi-object exploration. We evaluate our method on a set of object navigation tasks in both simulation as well as with a real robot, running in real-time on a Jetson Orin AGX. We demonstrate that it outperforms existing state-of-the-art approaches both on single and multi-object navigation tasks. Additional videos, code and the multi-object navigation benchmark will be available on \url{https://finnbsch.github.io/OneMap}.
%
\end{abstract}
%
%
%
\section{Introduction}

Object search in complex, cluttered environments presents significant challenges for robots, despite being relatively intuitive for humans. Humans can navigate novel environments by remembering semantic information along with the spatial layout of the area. As individuals explore new surroundings, they continuously update both of these while choosing search directions that semantically correlate with the object they are looking for, such as going to what looks like a kitchen when searching for the fridge. Additionally, by maintaining such an internal spatial-semantic representation over time we can also more efficiently guide our search for objects in the future. E.g., we may not have seen the oven, but we remember where the kitchen is. Here we want to similarly enable robots to remember such semantic-spatial information for multi-object search.

\begin{figure}[t]
    \centering
\centering
\def\svgwidth{\columnwidth}
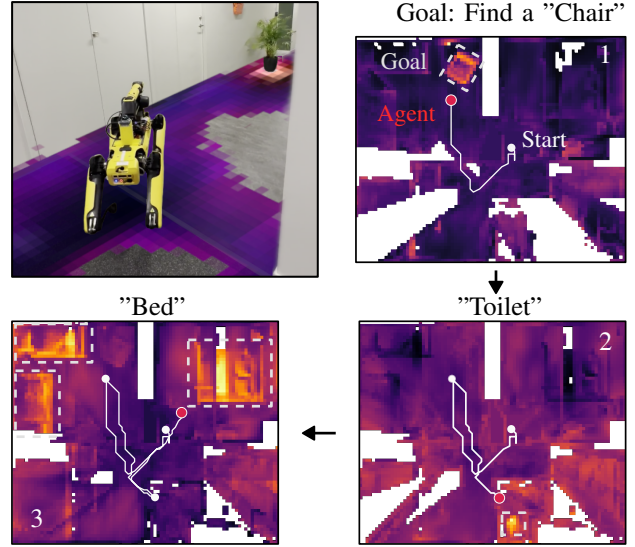
     \caption{Given a sequence of target object queries, e.g. (\textit{"chair"}, \textit{"toilet"}, \textit{"bed"}), OneMap enables an agent to perform efficient open-vocabulary multi-object navigation.}
    \label{Fig:placeholder2}
    \vspace{-0.5cm}
\end{figure}
Pre-trained Vision-Language Models (VLMs) have demonstrated the ability to correlate visual appearance and natural language, facilitating perceptual reasoning at increasingly abstract levels. Recent research has highlighted the benefits of using VLMs to inform zero-shot object navigation tasks~\cite{yokoyama2023vlfmvisionlanguagefrontiermaps}\cite{huang2023visuallanguagemapsrobot}.
However, there remains a gap in understanding of how to effectively leverage VLMs to inform search while maintaining reusable semantic-spatial information over consecutive object  navigation queries.

We propose a novel method to construct a real-time spatial-semantic map \textbf{(OneMap)} for efficient open-vocabulary object search. The contributions of our work include:
\begin{enumerate}
    \item We present an open-vocabulary, spatial-semantic map that is built and queried online, enabling multi-object navigation with retained semantic information across tasks. The method is developed for onboard deployment and we demonstrate it in real-world experiments running on a Jetson Orgin AGX.
    \item We propose a probabilistic observation model to mitigate common errors when mapping semantic features, and utilize the resulting probabilistic-semantic map to make uncertainty-informed decisions for semantic exploration.
    \item We introduce a new multi-object navigation benchmark to evaluate the semantic memory capabilities of autonomous agents, and release it as open source.
    \item We conduct extensive evaluations on benchmarks in the HM3D scene set \cite{ramakrishnan2021hm3d}. The results validate that our method outperforms state-of-the-art approaches both on single and multi-object navigation tasks.
\end{enumerate}
%
%
%
%
%
%
%
%
%
%
%
%

%
\section{Related Work}
%
%
%
%
%
%

Efforts in object goal navigation~\cite{batra2020objectnavrevisitedevaluationembodied} aim to equip mobile systems with the ability to use geometric and semantic properties of a given environment to efficiently guide an agent towards a target object. A large body of work has studied semantic goal navigation with a predefined set of object classes (c.f.~\cite{banerjee2022objectgoalnavigationbased}\cite{lin2024advancingobjectgoalnavigation}). Advances in large pre-trained vision-language models have further allowed the object navigation problem to be solved for objects and environments without prior training. In several recent works, such VLM-based approaches have been used for zero-shot object navigation.
%
%
%
\subsection{Zero-Shot Object Navigation}
Closest to our method is VLFM~\cite{yokoyama2023vlfmvisionlanguagefrontiermaps}. The method generates a 2D semantic similarity map, conditioned on an open-vocabulary text query, which is used to score frontiers when exploring the environment in search for a target object.
The authors propose to use the BLIP-2~\cite{li2023blip2bootstrappinglanguageimagepretraining} VLM to produce open-set, image-level features.
Instead, we extract patch-level CLIP-aligned features and formulate a probabilistic mapping scheme, which allows us to associate the features more accurately with locations in the map. Moreover, our method builds a map that contains language-queryable features, while VLFM only stores similarity scores for
a specific query.

The approach of \cite{zhou2023escexplorationsoftcommonsense} projects semantic object and room labels into a 2D navigation map with the help of the GLIP~\cite{li2022groundedlanguageimagepretraining} model, which yields language grounded bounding boxes. They additionally use a large language model, enabling commonsense reasoning about object locations. However, the integration of large language models typically bears the cost of high computational expense.
The works of~\cite{huang2023visuallanguagemapsrobot} and~\cite{chen2022openvocabularyqueryablescenerepresentations} present methods for constructing semantic navigation maps offline, which can be utilized during navigation.
\cite{huang2023visuallanguagemapsrobot} uses 3D projection of pixel-level features, extracted using LSeg~\cite{li2022languagedrivensemanticsegmentation}, which produces a highly detailed, queryable semantic representation.
In~\cite{chen2022openvocabularyqueryablescenerepresentations} the visual semantic information is combined with a large language model, which allows planning from unstructured natural language input. They follow~\cite{gu2022openvocabularyobjectdetectionvision} by combining a class agnostic region proposal network with CLIP~\cite{radford2021learningtransferablevisualmodels}.
\cite{Gadre2022CLIPOW} introduces and evaluates variations of a straightforward, heuristic frontier navigation method, using CLIP or open-set object detectors for guidance. 
%
%
%
%
\vspace{-0.1cm}
\subsection{Policy Learning for Goal Navigation}
In \cite{marza2023multi} and \cite{marza2022teaching} the authors propose a learning based approach with implicit neural semantic representations. The method allows reinforcement learning agents to find differently colored cylinders that were added to environments, but does not generalize to arbitrary objects. \cite{shah2022lmnavroboticnavigationlarge} proposes to combine an LLM, a VLM and a pre-trained visual navigation model~\cite{Shah_2021} to generate actionable plans from natural language inputs and a set of RGB observations. The authors of \cite{majumdar2023zsonzeroshotobjectgoalnavigation} propose to train navigation agents on RGB observations and semantic embeddings of goal images from the ImageNav~\cite{krantz2022instancespecificimagegoalnavigation} dataset. The work of~\cite{ramrakhya2023pirlnavpretrainingimitationrl} tackles the object-goal navigation problem via behavior cloning on human demonstrations with reinforcement learning based fine-tuning.
Generally, learned policies tend to exhibit specificity to the tasks and environments they were trained with.
%
%
%
\subsection{Open-Set Semantic 3D Mapping}

As compared to 2D navigation maps, 3D representations are usually more expensive in terms of memory and computation and typically not specifically designed for object goal navigation. Recently proposed 3D mapping approaches that store open-vocabulary semantic features include neural~\cite{qin2024langsplat3dlanguagegaussian}\cite{qiu2024learninggeneralizablefeaturefields}\cite{guo2024semanticgaussiansopenvocabularyscene} and non-neural metric-semantic representations~\cite{jatavallabhula2023conceptfusionopensetmultimodal3d}\cite{gu2023conceptgraphsopenvocabulary3dscene}\cite{yamazaki2023openfusionrealtimeopenvocabulary3d}\cite{takmaz2023openmask3dopenvocabulary3dinstance}. 
\cite{qin2024langsplat3dlanguagegaussian} and \cite{guo2024semanticgaussiansopenvocabularyscene} use semantic feature extraction in combination with gaussian splatting~\cite{kerbl20233d}, while~\cite{qiu2024learninggeneralizablefeaturefields} uses a generalizable NERF approach~\cite{mildenhall2021nerf} to ground semantics in implicit 3D representations.
The works of \cite{jatavallabhula2023conceptfusionopensetmultimodal3d}\cite{gu2023conceptgraphsopenvocabulary3dscene}\cite{takmaz2023openmask3dopenvocabulary3dinstance} extract semantic features using a combination of a segmentation model (SAM~\cite{kirillov2023segment}) and CLIP~\cite{radford2021learningtransferablevisualmodels} and project the features to 3D pointclouds.
In~\cite{yamazaki2023openfusionrealtimeopenvocabulary3d} the authors introduce a real-time volumetric representation using a region-level VLM~\cite{zou2023segment} at a small cost of accuracy. 
\section{Method}
\begin{figure*}[t]
\centering
\def\svgwidth{\textwidth}
\vspace{0.2cm}
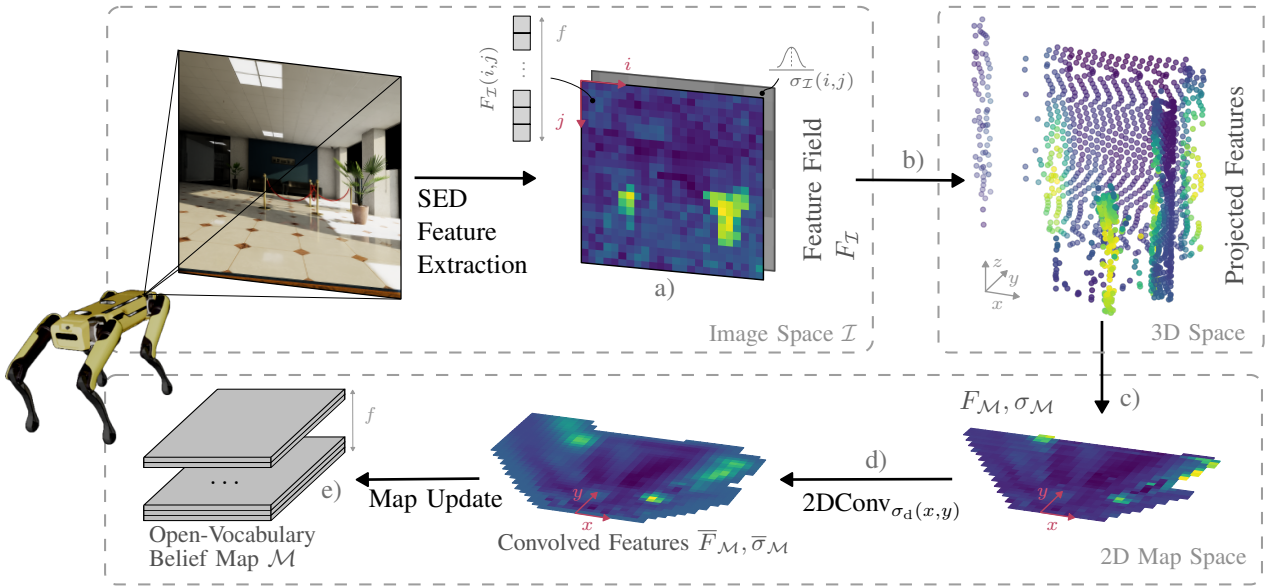
\caption{Overview of the map construction. We obtain a dense feature field with corresponding variances (a) from RGB images using the SED encoder~\cite{xie2024sedsimpleencoderdecoderopenvocabulary}, and project the features to 3D (b), and then to the 2D map space (c). We apply a sparse inverse Gaussian convolution (d) to account for uncertainties in the feature locations. Lastly, we update the open-vocabulary belief map (e) with the mapped features and obtain a dense, open-vocabulary queryable belief map.}
\label{fig:system_overview}
\vspace{-0.55cm}
\end{figure*}
%
This section describes our proposed method for real-time generation and exploration of open-set semantic belief maps via a probabilistic observation model of semantic features from a VLM. See Fig.~\ref{fig:system_overview} for a system overview. 
%
\subsection{Problem Formulation}
From a stream of RGB and depth images with associated camera poses we construct a two-dimensional, open-vocabulary belief map $\mathcal{M} = (F, \sigma^2)$ over CLIP-aligned features $F(x, y) \in \mathbb{R}^{n_x \times n_y \times f}$ and a variance per map location $\sigma^2(x, y) \in \mathbb{R}^{n_x \times n_y}$, where $f$ is the CLIP feature dimension.
We index the map $\mathcal{M}$ with $x \in [0, n_x], y \in [0, n_y]$ and data points in image-space with $i \in [0, H], j \in [0, W]$, where $n_x$, $n_y$ are the map's dimensions in grid cells and $H$, $W$ are the image height and width respectively in pixels. The memory requirements grow linearly with the number of observed cells. For instance, a map of $50\,\mathrm{m} \times 50\,\mathrm{m}$ with 10 cells per meter requires roughly $700\,\mathrm{Mb}$ of memory.

Given an RGB image $\mathcal{I} \in \mathbb{R}^{H \times W \times 3}$ in Fig.~\ref{fig:system_overview}a, we use SED~\cite{xie2024sedsimpleencoderdecoderopenvocabulary} to obtain a per-patch feature field $\mathcal{F} \in  \mathbb{R}^{H_\mathcal{F} \times W_\mathcal{F} \times {f}}$, where $H_\mathcal{F}$ and $W_\mathcal{F}$ are the image dimensions in patches.
The feature field is projected to 3D using the corresponding depth image $\mathcal{D} \in \mathbb{R}^{H \times W}$ and subsequently fused into the map in a probabilistic fashion, accounting for uncertainties in feature extraction and depth sensing.

The following sections describe the construction of the open-vocabulary belief map $\mathcal{M}(x,y)$ and detail how it is leveraged for effective object-goal navigation.
\subsection{Dense Semantic Feature Extraction}
We extract patch-level descriptive features in the CLIP~\cite{radford2021learningtransferablevisualmodels} feature space by employing the image encoder of SED~\cite{xie2024sedsimpleencoderdecoderopenvocabulary}. SED uses a hierarchical encoder based on the ConvNeXT~\cite{liu2022convnet2020s} architecture. This is motivated by the fact that ConvNeXT has linear complexity with respect to the size of the input data and has been shown to perform  well at capturing the locality of information when compared to plain transformers~\cite{xie2024sedsimpleencoderdecoderopenvocabulary}.
In SED, CLIP's pre-trained text encoder is frozen while the pipeline is trained to perform semantic segmentation, which produces a strong link between the patch locations and CLIP's multimodal feature space.

Note that the use of this feature extractor stands in contrast to other methods that use less accurate segmentation-level, or image-level features \cite{yokoyama2023vlfmvisionlanguagefrontiermaps}\cite{zhou2023escexplorationsoftcommonsense}, and allows us to construct more accurate, open-vocabulary maps in real-time. 
\subsection{Open-Vocabulary Belief Map Update}
To fuse the extracted semantic features into $\mathcal{M}(x,y)$, we employ a probabilistic observation model. First, we upsample the patch-level feature field $\mathcal{F}$ to match the dimensions of the image $\mathcal{I}$ using bilinear interpolation and assign each pixel $(i,j)$ of $\mathcal{I}$ the corresponding feature $F_\mathcal{I}(i,j)$. We model three sources of observation uncertainty, assuming independent, zero-mean normal noise distributions:
\begin{itemize}
    \item \textbf{Feature Extractor}: The reliability of semantic features varies depending on the distance to the camera. If an object is too far away, the image resolution might not be sufficient to obtain informative features. Conversely, if the camera is too close, the resulting image might lack the scene context needed to disambiguate what is in it.
    \item \textbf{Projective Feature Patch Leakage}: Large gradients in the depth image indicate sharp changes in environment geometry, e.g. a transition from an object in the foreground to an object in the background. In those areas, we assume the feature patches to be less trustworthy as features that are primarily influenced by the foreground might \textit{leak} to the background and vice versa.
    \item \textbf{Feature Location Uncertainty}: The accuracy of depth cameras decreases with distance, resulting in less reliable feature projections.
\end{itemize}
%
We incorporate the uncertainties stemming from \textit{Feature Leakage} and \textit{Feature Extraction} by assigning a variance value to pixels of $\mathcal{I}(i,j)$ (Fig.~\ref{fig:system_overview}a):
\begin{align}
    \sigma_\mathcal{I}^2 (i, j) = \sigma^2_\mathrm{L}(i, j)  \sigma^2_\mathrm{F}(i, j).
    \label{eq:totalvariance}
\end{align}
We model the feature leakage variance as
\begin{align}
    \sigma^2_\mathrm{L}(i,j) = \mathrm{tanh}(\nabla_x \mathcal{D}(i, j)^2 + \nabla_y \mathcal{D}(i, j)^2 ),
    \label{eq:gradvariance}
\end{align}
i.e. the standard deviation is linear in the gradient of the depth image. As a result, the system considers features less reliable where the depth image has large gradients. 

We assume the extracted features to be most reliable at an \textit{optimal detection distance} $d_{\mathrm{opt}}$, with their variance rapidly increasing for pixels closer or further to the camera.
The resulting variance is then given by
\begin{align}
    \sigma^2_\mathrm{F} (i, j) = \exp\left(\frac{(d_\mathrm{opt} - \mathcal{D}(i, j))}{2}^2\right),
    \label{eq:featurevariance}
\end{align}
where $d_{\mathrm{opt}}$ is tuned to match the model's performance.

Subsequently, we project each point $(i, j)$ to a 3D pointcloud of uncertain features $\mathcal{P}$ using $\mathcal{D}(i,j)$ and the camera intrinsics (Fig.~\ref{fig:system_overview}b). $\mathcal{P}$ is then projected onto the 2D plane. Features $F_\mathcal{I}(i,j)$ of points that come to lie in the same map cell are combined through a weighted summation approach, with weights $\propto\sigma_\mathcal{I}^2 (i, j)$, and the variances $\sigma_\mathcal{I}^2 (i, j)$ are averaged. As a result, we obtain a feature vector $F_\mathcal{M}(x, y)$ with corresponding variance $\sigma_\mathcal{M}^2 (x, y)$ for each map cell affected by the current observation, as seen in Fig.~\ref{fig:system_overview}c.

Lastly, we incorporate the uncertain locations of projected features, resulting from depth sensing noise, by blurring the features proportional to the location uncertainty. For this, we adopt the stereo camera accuracy model from \cite{grunnet2019subpixel} and model the variance of feature locations as quadratically dependent on the distance $d(x, y)$ between the map position and the camera:
\begin{align}
    \sigma^2_\mathrm{d}(x, y) = pd^2(x, y),
    \label{eq:depthvariance}
\end{align}
with $p$ a proportionality factor.

We achieve the blurring by convolving each mapped feature $F_\mathcal{M}(x, y)$ with a spatially varying Gaussian blur kernel.
The Gaussian convolution
\begin{align}
    \overline{F}_\mathcal{M}, \,\, \overline{\sigma}_{\mathcal{M}}^2 = \mathrm{2DConv}_{\sigma^2_\mathrm{d}(x, y)}(F_\mathcal{M}, \sigma^2_\mathcal{M})
\end{align}
is parameterized by the spatially-varying $\sigma^2_\mathrm{d}(x, y)$ and affects both, features and variances. We obtain the convolved features $\overline{F}_\mathcal{M}(x,y)$ and variances $\overline{\sigma}_{\mathcal{M}}^2(x,y)$ in map space.
Intuitively, this results in stronger blur for distant geometries, accounting for the loss of accuracy in the spatial grounding of the semantics.
Conversely, the scene is resolved in higher detail if the camera is close, as visualized in Fig.~\ref{fig:system_overview}d. To ensure real-time feasibility and limit memory usage on-board the robot, we implement this as a custom sparse inverse Gaussian convolution operation, optimized for GPU.

The convolved features are fused into the persistent map $\mathcal{M}$ per grid cell following the scheme of recursive Bayesian estimation (Fig.~\ref{fig:system_overview}e), using the Kalman gain $K$:
%
%
%
\begin{align}
    K &= \frac{{\sigma_t}^2(x, y)}{\overline{\sigma}_\mathcal{M}^2(x, y) + \sigma_t^2(x, y)}, 
    \label{Eq:bayes1}\\[2pt]
    F_{t+1}(x, y) &= F_t(x, y) + K (\overline{F}_\mathcal{M} (x, y) - F_t(x, y)),
    \label{Eq:bayes2}\\[1pt]
    \sigma^2_{t + 1}(x, y) &= (1 - K) \sigma^2_{t}(x, y).
    \label{Eq:bayes3}
\end{align}
\subsection{Object Search}
\label{seq:nav}
\begin{figure}[t]
\centering
\def\svgwidth{0.65\columnwidth}
\vspace{0.1cm}
\begingroup%
  \makeatletter%
  \providecommand\color[2][]{%
    \errmessage{(Inkscape) Color is used for the text in Inkscape, but the package 'color.sty' is not loaded}%
    \renewcommand\color[2][]{}%
  }%
  \providecommand\transparent[1]{%
    \errmessage{(Inkscape) Transparency is used (non-zero) for the text in Inkscape, but the package 'transparent.sty' is not loaded}%
    \renewcommand\transparent[1]{}%
  }%
  \providecommand\rotatebox[2]{#2}%
  \newcommand*\fsize{\dimexpr\f@size pt\relax}%
  \newcommand*\lineheight[1]{\fontsize{\fsize}{#1\fsize}\selectfont}%
  \ifx\svgwidth\undefined%
    \setlength{\unitlength}{299.56983191bp}%
    \ifx\svgscale\undefined%
      \relax%
    \else%
      \setlength{\unitlength}{\unitlength * \real{\svgscale}}%
    \fi%
  \else%
    \setlength{\unitlength}{\svgwidth}%
  \fi%
  \global\let\svgwidth\undefined%
  \global\let\svgscale\undefined%
  \makeatother%
  \begin{picture}(1,0.58008886)%
    \lineheight{1}%
    \setlength\tabcolsep{0pt}%
    \put(0,0){\includegraphics[width=\unitlength,page=1]{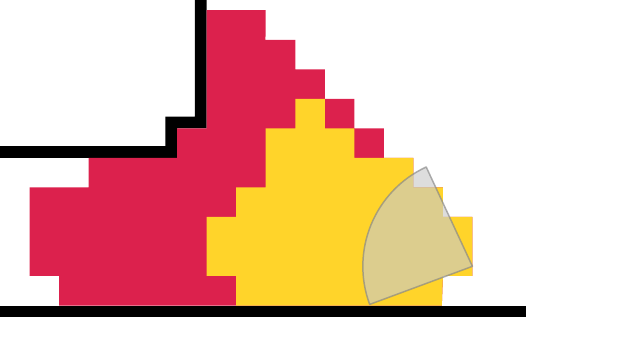}}%
    \put(0.9465346,0.19927285){\color[rgb]{0.35294118,0.35294118,0.35294118}\makebox(0,0)[t]{\lineheight{1.25}\smash{\begin{tabular}[t]{c}Agent\end{tabular}}}}%
    \put(0.73760017,0.40608346){\color[rgb]{0,0,0}\makebox(0,0)[t]{\lineheight{1.25}\smash{\begin{tabular}[t]{c}$\sim \mathcal{N}$\end{tabular}}}}%
    \put(0,0){\includegraphics[width=\unitlength,page=2]{navmap.pdf}}%
    \put(0.19650396,0.21508657){\color[rgb]{0,0,0}\makebox(0,0)[t]{\lineheight{1.25}\smash{\begin{tabular}[t]{c}$\mathcal{O}$\end{tabular}}}}%
    \put(0.5677432,0.21508657){\color[rgb]{0,0,0}\makebox(0,0)[t]{\lineheight{1.25}\smash{\begin{tabular}[t]{c}$\mathcal{E}$\end{tabular}}}}%
    \put(0,0){\includegraphics[width=\unitlength,page=3]{navmap.pdf}}%
    \put(0.33118353,0.00539679){\color[rgb]{0,0.39215686,0.87058824}\makebox(0,0)[t]{\lineheight{1.25}\smash{\begin{tabular}[t]{c}$\delta$\end{tabular}}}}%
    \put(0,0){\includegraphics[width=\unitlength,page=4]{navmap.pdf}}%
  \end{picture}%
\endgroup%

\caption{The observed ($\mathcal{O}$), fully explored ($\mathcal{E}$), and non-navigable ($\sim \mathcal{N}$) map with an agent venturing into unknown areas. The blue line $\delta$ marks the frontier between $\mathcal{E}$ and $\mathcal{O}$.}
\label{fig:navmap}
\vspace{-0.55cm}
\end{figure}

%
For a given text prompt we use the language encoder of CLIP~\cite{radford2021learningtransferablevisualmodels} to produce an embedded object query $Q \in \mathbb{R}^f$.
We calculate the cosine similarity between $Q$ and each map cell to obtain a query-conditioned similarity map $\mathcal{S}_Q(x, y) \in \mathbb{R}^{n_x \times n_y}$. For the purpose of object exploration we keep track of an additional variance estimate $\sigma_\mathrm{E}(x,y)$, which is analogously updated according to Eq.~\ref{Eq:bayes3}. In contrast to $\sigma(x, y)$, $\sigma_\mathrm{E}(x,y)$ can be reset during the exploration process, as laid out in the following. We introduce the notion of four binary sub-maps, shown in Fig.~\ref{fig:navmap}:
\begin{itemize}
    \item \textbf{Observed Map} $\mathcal{O}$: The part of the map that has been observed by any update.
    \item \textbf{Semantically Explored Map} $\mathcal{E}$: Derived by thresholding $\sigma(x,y)$. It represents the area we consider \textit{semantically explored}, i.e. we assume that the mapped features are spatially accurate and sufficiently descriptive of the environment. It is a subset of the observed map.
    \item \textbf{Searched Map} $\mathcal{C}$: Derived by thresholding $\sigma_\mathrm{E}(x,y)$. It represents the area we consider explored in the context of the current object search query $Q_i$. Once the system receives a new query $Q_{i+1}$, then $\sigma_\mathrm{E}(x,y)$ is reset.
    \item \textbf{Navigable Map} $\mathcal{N}$: Subset of the map that is free of obstacles, used by the path planner.
\end{itemize}

With the help of these sub-maps we compute two types of navigation goals: Firstly we compute frontiers $\delta$ on the border between the semantically explored and the observed map, shown in Fig.~\ref{fig:navmap}. We score them based on the highest query-to-feature similarity $\mathrm{max}(\mathcal{S}_Q(x, y))$ in map cells in $\mathcal{O} - \mathcal{E}$ reachable from the frontier, i.e. cells that are observed but not semantically explored.

Secondly we cluster regions with high values of $\mathcal{S}_Q(x, y)$ in the semantically explored, but not searched map, $\mathcal{E} - \mathcal{C}$. Similarly, we score them with respect to the maximum similarity value contained in the cells of a cluster.

Note that the latter type of navigation goals are only obtained if the searched map is different from the semantically explored map, i.e. the searched map has been reset at least once.
Hence, upon obtaining a new text prompt, e.g. after successfully navigating to a goal, $\sigma_\mathrm{E}(x,y)$ is reset, allowing previously seen high-similarity areas to be eligible as navigation targets again.

We then greedily select the highest-scoring navigation goal and use $A^*$~\cite{hart1968astar} to plan a collision free path.
We further employ a visual object detector as a second measure to identify a potential target object. We apply consensus-filtering between the map and object detector by only considering an object as detected when 1) the object detector triggers, and 2) the corresponding area in the map $\mathcal{S}_Q(x, y)$ has a similarity score within the top 5-percentile. Upon detection, the agent navigates close to the object and terminates the exploration process. We follow~\cite{yokoyama2023vlfmvisionlanguagefrontiermaps} and employ YOLOv7~\cite{wang2023yolov7} if the object is
part of the classes in MS-COCO~\cite{lin2015microsoftcococommonobjects} and open-set detector Yolo-World~\cite{Cheng2024YOLOWorld} otherwise.
\vspace{-0.1cm}
\section{Experimental Setup}
\label{seq:expset}
\vspace{-0.1cm}
We evaluate our method on a standard single-object simulation navigation benchmark, on a novel benchmark that we construct for the multi-object navigation problem, as well as on navigation experiments with a real robot. 
\\\textbf{Navigation Tasks.} 
\textit{Single-Object Navigation}~\cite{batra2020objectnavrevisitedevaluationembodied} requires the agent to navigate from a given initial position to any instance of a target object category.
We evaluate on the validation split of the Habitat ObjectNav Challenge~\cite{habitatchallenge2022} based on the Habitat Matterport 3D~(HM3D)~\cite{ramakrishnan2021hm3d} dataset. It comprises 2000 single-object episodes in 20 scenes over six object categories. 

\textit{Multi-Object Navigation:} We define multi-object navigation as a sequence of object goals. The agent is asked to navigate to a given object category. Upon successfully finding the object, the agent is informed about the next object goal. The episode concludes if the agent does not succeed in finding a given object.
Goal navigation is considered successful if the agent terminates the task and if it terminates in a position within $1.5\,\mathrm{m}$ of any instance of the target object category.
To construct the task, we build on the Habitat ObjectNav challenge: We generate episodes by sampling from starting poses present in the single-object task and then randomly generate a sequence of reachable objects, that are reachable without traversing stairs.
Our dataset consists of 236 episodes in 20 scenes over six object categories. Each episode consists of a sequence of three goal objects. The dataset generation is further detailed on our project page.

\textbf{Metrics.} For single-object navigation, we report the Success Rate (SR) and Success weighed by Path Length (SPL)~\cite{anderson2018evaluationembodiednavigationagents}. For successful episodes, the SPL is the optimal path length divided by the agent's path length, else zero.

For multi-object navigation, we report the overall success rate (SR), the fraction of episodes in which all objects were found, as well as the overall success weighted by path length (SPL). The SPL is the optimal path length divided by the agent's path length if all objects were found, and zero otherwise. The optimal path is hereby defined as the sum of \textit{sequence-wise} optimal paths, i.e. the shortest path to the next object. This is done to account for the fact that the agent gets informed about the next target object only after succeeding in finding the previous one.
Moreover, following~\cite{wani2020multion}, we report the progress (PR) and the progress weighed by path length (PPL). The progress is the fraction of found objects per episode. Note that an episode is terminated if the agent fails to find or misidentifies any target object of the episode.

\textbf{Baselines.} For single-object navigation, we compare against two SOTA zero-shot methods: ESC~\cite{zhou2023escexplorationsoftcommonsense}, and VLFM~\cite{yokoyama2023vlfmvisionlanguagefrontiermaps}. Similar to our method, ESC and VLFM perform frontier-based exploration. ESC scores frontiers based on object detections in proximity of the frontier by querying an LLM and VLFM queries BLIP2 to determine promising frontiers. Moreover, we compare against two methods that either require task-specific training or supervision.
ZSON~\cite{majumdar2023zsonzeroshotobjectgoalnavigation} employs CLIP to transfer a model trained for ImageNav to the ObjectNav task. PIRLNav~\cite{ramrakhya2023pirlnavpretrainingimitationrl} is an end-to-end policy trained on human demonstrations.

To the best of our knowledge, ours is the first method that performs real-time, zero-shot multi-object navigation and without extensive task and environment-specific training.
Hence, for multi-object navigation, we compare against VLFM as the strongest performing baseline for the single-object task. 
\vspace{-0.3cm}
\section{Results}
\vspace{-0.1cm}
%
%
Our experiments are designed to answer the following questions: (1) How does our method compare against state of the art methods for single-object navigation? 
(2) Can our method effectively use its semantic map to improve performance over a sequence of multiple object goals compared to existing approaches? (3) Can our method be successfully be deployed on real robot searching real environments?
%
\subsection{Single-Object Navigation}
\label{seq:son}
\begin{figure}
\centering
\def\svgwidth{0.9\columnwidth}
\vspace{0.23cm}
\begingroup%
  \makeatletter%
  \providecommand\color[2][]{%
    \errmessage{(Inkscape) Color is used for the text in Inkscape, but the package 'color.sty' is not loaded}%
    \renewcommand\color[2][]{}%
  }%
  \providecommand\transparent[1]{%
    \errmessage{(Inkscape) Transparency is used (non-zero) for the text in Inkscape, but the package 'transparent.sty' is not loaded}%
    \renewcommand\transparent[1]{}%
  }%
  \providecommand\rotatebox[2]{#2}%
  \newcommand*\fsize{\dimexpr\f@size pt\relax}%
  \newcommand*\lineheight[1]{\fontsize{\fsize}{#1\fsize}\selectfont}%
  \ifx\svgwidth\undefined%
    \setlength{\unitlength}{256.04433555bp}%
    \ifx\svgscale\undefined%
      \relax%
    \else%
      \setlength{\unitlength}{\unitlength * \real{\svgscale}}%
    \fi%
  \else%
    \setlength{\unitlength}{\svgwidth}%
  \fi%
  \global\let\svgwidth\undefined%
  \global\let\svgscale\undefined%
  \makeatother%
  \begin{picture}(1,0.4110334)%
    \lineheight{1}%
    \setlength\tabcolsep{0pt}%
    \put(0,0){\includegraphics[width=\unitlength,page=1]{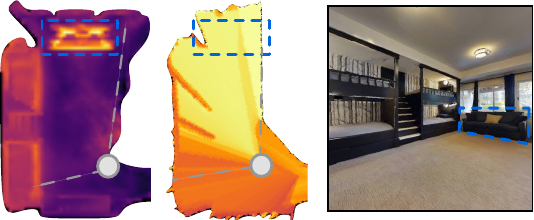}}%
    \put(0.15675325,0.39322944){\color[rgb]{0,0.39215686,0.87058824}\makebox(0,0)[lt]{\lineheight{1.25}\smash{\begin{tabular}[t]{l}\small Target Object\end{tabular}}}}%
    \put(0.22145838,0.12026348){\color[rgb]{0.58823529,0.58823529,0.58823529}\makebox(0,0)[lt]{\lineheight{1.25}\smash{\begin{tabular}[t]{l}\small Agent\end{tabular}}}}%
    \put(0,0){\includegraphics[width=\unitlength,page=2]{mapcom.pdf}}%
  \end{picture}%
\endgroup%
\caption{Comparison of our map (left) vs VLFM's map (right) taken from an episode of the HM3D dataset as the agent enters a new room, with the corresponding RGB observation. VLFM's \textit{similarity} map built for the target object \textit{couch} yields less spatially accurate similarity scores than our \textit{feature} map queried for the same.}%
\label{fig:mapcomp}%
\vspace{-0.4cm}%
\end{figure}

%
%
%
%
We compare our approach against all baseline methods on the single object navigation dataset, detailed in Section~\ref{seq:expset}. The corresponding results are presented in Table~\ref{tab:single_object_res}. Our method outperforms the state-of-the-art zero-shot methods, VLFM and ESC, in both success weighed by path length (SPL) and success rate (SR). Specifically, it achieves an SPL of $37.4\%$ and a success rate of $55.8\%$. When comparing with methods that require task- and environment-specific training, our method OneMap still outperforms the baselines PIRLNav and ZSON in terms of SPL, surpassing them by $10.3\%$ and $24.8\%$, respectively. Although PIRLNav demonstrates a $8.3\%$ higher success rate, it requires additional time-consuming training and human demonstrations (77k demonstrations, accounting for ${\sim 2378}$ human annotation hours) for the specific task and environment, whereas our method achieves competitive results zero-shot.
\begin{table}[h]
\centering
\vspace{-0.1cm}
\begin{tabular}{lcccc}
\toprule
\multirow{2}{*}{Approach} & \multicolumn{2}{c}{Training} & \multicolumn{2}{c}{HM3D} \\
\cmidrule(lr){2-3} \cmidrule(lr){4-5}
 & Locom. & Sem. & SPL$\uparrow$ & SR$\uparrow$ \\
\midrule
PIRLNav~\cite{ramrakhya2023pirlnavpretrainingimitationrl} & \cmark & \cmark & 27.1 & \textbf{64.1} \\
ZSON~\cite{majumdar2023zsonzeroshotobjectgoalnavigation} & \cmark & \cmark \tablefootnote{Trained for ImageNav~\cite{krantz2022instancespecificimagegoalnavigation}, and transferred to ObjectNav using CLIP.} & 12.6 & 25.5 \\
\midrule
ESC~\cite{zhou2023escexplorationsoftcommonsense} & \xmark & \xmark & 22.3 & 39.2 \\
VLFM~\cite{yokoyama2023vlfmvisionlanguagefrontiermaps}& \cmark & \xmark & 30.4 & 52.5 \\
OneMap (Ours) & \xmark & \xmark & \textbf{37.4} & \textbf{55.8} \\
\bottomrule
\end{tabular}
\caption{Single-Object Navigation Results in HM3D. We compare against two methods that require task/environment training  (upper rows), and two zero-shot methods like ours (lower rows). Our method outperforms all other methods in terms of success rate weighted by path length (SPL).}
\label{tab:single_object_res}
\vspace{-0.3cm}
\end{table}
Our superior performance in SPL indicates that our  proposed semantic feature map provides more informative guidance, resulting in shorter paths to the target object. However, it is interesting that we also outperform VLFM on success rate alone, as this does not factor in path length and we use the same object detector. Moreover, VLFM uses a PointNav locomotion policy trained specifically to better traverse the HM3D domain while we simply compute the shortest path to the frontier on a grid. We suspect that our higher SR compared to VLFM on the single-target task can be attributed to our ability to do consensus-filtering with our high-confidence semantic beliefs as outlined in Section~\ref{seq:nav}, whereas VLFM only uses heuristics to mitigate false detections. 3D reconstruction datasets such as HM3D additionally suffer from artifacts \cite{ramakrishnan2021hm3d} which can exacerbate misdetections. To investigate our surprisingly good results on single-object SR further, we repeat the evaluation without any consensus-filtering or heuristics to mitigate false detections at all. Instead, we just accept all detections from the detector. The results of this ablation shown in Table~\ref{tab:misdetect_ablation} indicate that false positives are a considerable source of error on the ObjectNav task and our filtering can effectively reduce the false positive rate (FP) of the object detector by $7.1\%$. We note that we still achieve higher SPL than VLFM (with its misdetection heuristics and domain-specific locomotion policy). We attribute this to our semantic feature map simply being more informative. Note that our consensus-filtering requires spatially accurate features, which our map provides, as opposed to VLFM, as shown in Fig.~\ref{fig:mapcomp}, and hence would not yield the same benefit if applied to less accurate maps. 
Overall, we conclude that our semantic belief map can more effectively guide an agent for single-object navigation tasks, and moreover it also reduces false positives from the  object detection stage.
\begin{table}[h]
\vspace{-0.2cm}%
\centering
\begin{tabularx}{0.78\columnwidth}{XXXX}
\toprule
{} & FP $\downarrow$ & SPL$\uparrow$ & SR$\uparrow$ \\
\midrule
With CF & \textbf{19.5} & \textbf{37.4} & \textbf{55.8} \\
W/o CF & 26.6 & 31.2 & 48.2 \\
\bottomrule
\end{tabularx}
\caption{Single-Object metrics and false positive rate (FP) for our method with vs. without consensus-filtering (CF)}
\label{tab:misdetect_ablation}
\vspace{-0.3cm}
\end{table}
\subsection{Multi-Object Navigation}
\label{seq:mon}
\begin{figure}
\centering
{%
\small%
\vspace{0.15cm}
\begin{tikzpicture}
    \begin{axis}[
        width=1.0\columnwidth,
        height=0.5\columnwidth,
        ybar=5pt,
        bar width=14pt,
        enlarge x limits=0.35,
        legend style={
            at={(0.02,0.95)}, 
            anchor=north west, 
            legend columns=-1, 
            font=\small,
            /tikz/every even column/.append style={column sep=0.5cm}
        },
        legend image code/.code={
            \draw[#1, fill=#1] (0cm,-0.1cm) rectangle (0.2cm,0.1cm);
        },
        ylabel={SPL},
        xlabel={\small Object Number},
        xtick={1,2,3},
        xticklabels={1,2,3},
        nodes near coords,
        nodes near coords align={vertical},
        x label style={at={(axis description cs:0.5,-0.17)},anchor=north},
        ymin=25,
        ymax=50,
        ymajorgrids=true,
        xmajorgrids=true,
        grid style={lightgrey, dashed},
    ]%
    \addplot[fill=mumblue] coordinates {(1,27.50) (2,32.61) (3,29.13)};%
    \addplot[fill=mumred] coordinates {(1,35.08) (2,37.44) (3,46.01)};%
    \legend{VLFM, OneMap}%
    \end{axis}%
\end{tikzpicture}%
%
}%
\vspace{-0.25cm}
\caption{SPL for the first, second, and third object goal. Our method effectively uses the information stored in the map, as evident in improved performance for later object goals.}%
\label{fig:results_per_seq}%
\vspace{-0.6cm}
\end{figure}
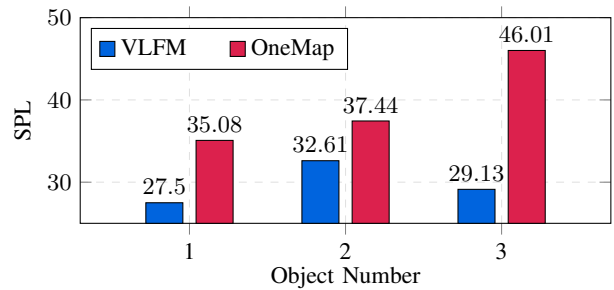

\begin{table}
\vspace{-0.5cm}%
\centering
\begin{tabular}{lcccccc}
\toprule
\multirow{2}{*}{Approach} & \multicolumn{2}{c}{Training} & \multicolumn{4}{c}{HM3D} \\
\cmidrule(lr){2-3} \cmidrule(lr){4-7}
 & Locom. & Sem.  & SPL$\uparrow$ & SR$\uparrow$ & PPL$\uparrow$ & PR$\uparrow$ \\
\midrule
VLFM~\cite{yokoyama2023vlfmvisionlanguagefrontiermaps} & \cmark & \xmark & 19.90 & 44.92 & 24.87 & 57.34\\
OneMap (Ours) & \xmark & \xmark & \textbf{27.77} & \textbf{54.24} &  \textbf{32.68} & \textbf{65.54} \\
\bottomrule
\end{tabular}%
\caption{Multi-Object Navigation Results in HM3D, compared against VLFM.}%
\label{tab:multi_obj_results}
\end{table}
We evaluate the performance of our method OneMap against VLFM on the multi-object navigation task in HM3D. Table~\ref{tab:multi_obj_results} highlights our method's superiority over the baseline VLFM across all metrics. We achieve an average progress (PR) of $65.54\%$, meaning that we successfully find 2 objects per episode on average. Moreover, as expected the performance gap to VLFM grows larger for multi-object tasks than for single-object tasks, indicating that our method can effectively reuse the map for multiple objects.
We evaluate this further in Fig.~\ref{fig:results_per_seq}, where we compute the average SPL for the first, second, and third object goals of each episode, given that the previous object was successfully found. As evident in the figure, the efficiency (SPL) for our method monotonically increases, whereas it remains relatively constant for VLFM. This can be attributed to our uncertainty-aware exploration resulting in high-quality open-vocabulary navigation maps that can effectively reuse the data from previous object goals, unlike VLFM, which builds a query-specific scoring map that cannot leverage prior searches. As a result, our method sees a relative  increase in efficiency (SPL) by a factor of 1.5x compared to VLFM for the third object, benefiting from the high-quality open-vocabulary semantic map constructed during the search for the first and second objects. We note that $100\%$ SPL here represents perfect (oracle) knowledge of where every object is in advance and is therefore an unattainable upper bound.
%
%
%
%
\vspace{-0.02cm}
\subsection{Real World Experiments}
\label{seq:rwe}
We deploy OneMap on a Boston Dynamics Spot quadruped using only one front-facing Realsense D455 stereo depth camera, and a Livox Mid 360 lidar with Fast-LIO2~\cite{xu2021fastlio2fastdirectlidarinertial} for odometry. We use Yolo-World~\cite{Cheng2024YOLOWorld} instead of YOLOv7 to allow searching arbitrary objects, instead of MS-COCO~\cite{lin2015microsoftcococommonobjects} classes only.
We run the complete system on-board a Jetson Orin AGX, and achieve an overall update frequency of $2\,\mathrm{Hz}$, which we found to be sufficient for an indoor, legged robot. 

For deploying our method in the real-world, we only adapt the feature localization parameter in Eq.~\ref{eq:depthvariance} to account for larger depth noise, but conduct no additional tuning of method parameters used for the evaluation in simulation. We follow the planned paths with a path-tracking controller and send velocity commands to the robot. Real-world experiments of multi-object navigation tasks can be viewed at \url{https://finnbsch.github.io/OneMap}.
\vspace{-0.1cm}
\section{Conclusion}
In this paper we present a real-time capable, zero-shot, semantic object navigation method that builds a reusable semantic belief map from patch-level language-aligned semantic image features. Leveraging the rich visual-semantic space of CLIP~\cite{radford2021learningtransferablevisualmodels} and performing a probabilistic map update is shown to yield a high-quality semantic representation that provides state-of-the-art performance on single-object as well as multi-object open-vocabulary navigation tasks. Furthermore we demonstrate the method's real-world applicability via successful experiments running onboard a quadruped robot. 

We found that our method works well with projection to 2D, but future work could include examining suitable 3D map representations. The presented mapping method can readily be adapted to produce a 3D voxel map, though this will introduce additional complexity to the mapping and planning parts. Moreover, means to address noise stemming from the reliance on depth images should be addressed. Lastly, the applicability of the method to outdoor scenarios where semantics might be less meaningful needs to be studied.

%
\addtolength{\textheight}{-0cm}   

\bibliographystyle{IEEEtran}
\bibliography{bib}

\begin{thebibliography}{10}
\providecommand{\url}[1]{#1}
\csname url@samestyle\endcsname
\providecommand{\newblock}{\relax}
\providecommand{\bibinfo}[2]{#2}
\providecommand{\BIBentrySTDinterwordspacing}{\spaceskip=0pt\relax}
\providecommand{\BIBentryALTinterwordstretchfactor}{4}
\providecommand{\BIBentryALTinterwordspacing}{\spaceskip=\fontdimen2\font plus
\BIBentryALTinterwordstretchfactor\fontdimen3\font minus \fontdimen4\font\relax}
\providecommand{\BIBforeignlanguage}[2]{{%
\expandafter\ifx\csname l@#1\endcsname\relax
\typeout{** WARNING: IEEEtran.bst: No hyphenation pattern has been}%
\typeout{** loaded for the language `#1'. Using the pattern for}%
\typeout{** the default language instead.}%
\else
\language=\csname l@#1\endcsname
\fi
#2}}
\providecommand{\BIBdecl}{\relax}
\BIBdecl

\bibitem{yokoyama2023vlfmvisionlanguagefrontiermaps}
N.~Yokoyama, S.~Ha, D.~Batra, J.~Wang, and B.~Bucher, ``Vlfm: Vision-language frontier maps for zero-shot semantic navigation,'' in \emph{International Conference on Robotics and Automation (ICRA)}, 2024.

\bibitem{huang2023visuallanguagemapsrobot}
C.~Huang, O.~Mees, A.~Zeng, and W.~Burgard, ``Visual language maps for robot navigation,'' in \emph{Proceedings of the IEEE International Conference on Robotics and Automation (ICRA)}, London, UK, 2023.

\bibitem{ramakrishnan2021hm3d}
\BIBentryALTinterwordspacing
S.~K. Ramakrishnan, A.~Gokaslan, E.~Wijmans, O.~Maksymets, A.~Clegg, J.~M. Turner, E.~Undersander, W.~Galuba, A.~Westbury, A.~X. Chang, M.~Savva, Y.~Zhao, and D.~Batra, ``Habitat-matterport 3d dataset ({HM}3d): 1000 large-scale 3d environments for embodied {AI},'' in \emph{Thirty-fifth Conference on Neural Information Processing Systems Datasets and Benchmarks Track}, 2021. [Online]. Available: \url{https://arxiv.org/abs/2109.08238}
\BIBentrySTDinterwordspacing

\bibitem{batra2020objectnavrevisitedevaluationembodied}
\BIBentryALTinterwordspacing
D.~Batra, A.~Gokaslan, A.~Kembhavi, O.~Maksymets, R.~Mottaghi, M.~Savva, A.~Toshev, and E.~Wijmans, ``Objectnav revisited: On evaluation of embodied agents navigating to objects,'' 2020. [Online]. Available: \url{https://arxiv.org/abs/2006.13171}
\BIBentrySTDinterwordspacing

\bibitem{banerjee2022objectgoalnavigationbased}
D.~S. Chaplot, D.~Gandhi, A.~Gupta, and R.~Salakhutdinov, ``Object goal navigation using goal-oriented semantic exploration,'' in \emph{In Neural Information Processing Systems (NeurIPS)}, 2020.

\bibitem{lin2024advancingobjectgoalnavigation}
\BIBentryALTinterwordspacing
M.~Lin, Y.~Chen, D.~Zhao, and Z.~Wang, ``Advancing object goal navigation through llm-enhanced object affinities transfer,'' 2024. [Online]. Available: \url{https://arxiv.org/abs/2403.09971}
\BIBentrySTDinterwordspacing

\bibitem{li2023blip2bootstrappinglanguageimagepretraining}
J.~Li, D.~Li, S.~Savarese, and S.~Hoi, ``Blip-2: bootstrapping language-image pre-training with frozen image encoders and large language models,'' in \emph{Proceedings of the 40th International Conference on Machine Learning}, 2023, pp. 19\,730--19\,742.

\bibitem{zhou2023escexplorationsoftcommonsense}
K.~Zhou, K.~Zheng, C.~Pryor, Y.~Shen, H.~Jin, L.~Getoor, and X.~E. Wang, ``Esc: Exploration with soft commonsense constraints for zero-shot object navigation,'' in \emph{Proceedings of the 40th International Conference on Machine Learning}, ser. ICML'23.\hskip 1em plus 0.5em minus 0.4em\relax JMLR.org, 2023.

\bibitem{li2022groundedlanguageimagepretraining}
L.~H. Li, P.~Zhang, H.~Zhang, J.~Yang, C.~Li, Y.~Zhong, L.~Wang, L.~Yuan, L.~Zhang, J.-N. Hwang \emph{et~al.}, ``Grounded language-image pre-training,'' in \emph{2022 IEEE/CVF Conference on Computer Vision and Pattern Recognition (CVPR)}.\hskip 1em plus 0.5em minus 0.4em\relax IEEE, 2022, pp. 10\,955--10\,965.

\bibitem{chen2022openvocabularyqueryablescenerepresentations}
B.~Chen, F.~Xia, B.~Ichter, K.~Rao, K.~Gopalakrishnan, M.~S. Ryoo, A.~Stone, and D.~Kappler, ``Open-vocabulary queryable scene representations for real world planning,'' in \emph{2023 IEEE International Conference on Robotics and Automation (ICRA)}.\hskip 1em plus 0.5em minus 0.4em\relax IEEE, 2023, pp. 11\,509--11\,522.

\bibitem{li2022languagedrivensemanticsegmentation}
\BIBentryALTinterwordspacing
B.~Li, K.~Q. Weinberger, S.~Belongie, V.~Koltun, and R.~Ranftl, ``Language-driven semantic segmentation,'' in \emph{International Conference on Learning Representations}, 2022. [Online]. Available: \url{https://openreview.net/forum?id=RriDjddCLN}
\BIBentrySTDinterwordspacing

\bibitem{gu2022openvocabularyobjectdetectionvision}
X.~Gu, T.-Y. Lin, W.~Kuo, and Y.~Cui, ``Open-vocabulary object detection via vision and language knowledge distillation,'' in \emph{International Conference on Learning Representations}.

\bibitem{radford2021learningtransferablevisualmodels}
A.~Radford, J.~W. Kim, C.~Hallacy, A.~Ramesh, G.~Goh, S.~Agarwal, G.~Sastry, A.~Askell, P.~Mishkin, J.~Clark \emph{et~al.}, ``Learning transferable visual models from natural language supervision,'' in \emph{International conference on machine learning}.\hskip 1em plus 0.5em minus 0.4em\relax PMLR, 2021, pp. 8748--8763.

\bibitem{Gadre2022CLIPOW}
S.~Y. Gadre, M.~Wortsman, G.~Ilharco, L.~Schmidt, and S.~Song, ``Cows on pasture: Baselines and benchmarks for language-driven zero-shot object navigation,'' \emph{CVPR}, 2023.

\bibitem{marza2023multi}
P.~Marza, L.~Matignon, O.~Simonin, and C.~Wolf, ``Multi-object navigation with dynamically learned neural implicit representations,'' in \emph{International Conference on Computer Vision (ICCV)}, 2023.

\bibitem{marza2022teaching}
------, ``Teaching agents how to map: Spatial reasoning for multi-object navigation,'' \emph{International Conference on Intelligent Robots and Systems (IROS)}, 2022.

\bibitem{shah2022lmnavroboticnavigationlarge}
\BIBentryALTinterwordspacing
D.~Shah, B.~Osinski, B.~Ichter, and S.~Levine, ``{LM}-nav: Robotic navigation with large pre-trained models of language, vision, and action,'' in \emph{6th Annual Conference on Robot Learning}, 2022. [Online]. Available: \url{https://openreview.net/forum?id=UW5A3SweAH}
\BIBentrySTDinterwordspacing

\bibitem{Shah_2021}
\BIBentryALTinterwordspacing
D.~Shah, B.~Eysenbach, G.~Kahn, N.~Rhinehart, and S.~Levine, ``{ViNG: Learning Open-World Navigation with Visual Goals},'' in \emph{IEEE International Conference on Robotics and Automation (ICRA)}, 2021. [Online]. Available: \url{https://arxiv.org/abs/2012.09812}
\BIBentrySTDinterwordspacing

\bibitem{majumdar2023zsonzeroshotobjectgoalnavigation}
A.~Majumdar, G.~Aggarwal, B.~Devnani, J.~Hoffman, and D.~Batra, ``Zson: Zero-shot object-goal navigation using multimodal goal embeddings,'' in \emph{Neural Information Processing Systems (NeurIPS)}, 2022.

\bibitem{krantz2022instancespecificimagegoalnavigation}
J.~Krantz, S.~Lee, J.~Malik, D.~Batra, and D.~S. Chaplot, ``Instance-specific image goal navigation: Training embodied agents to find object instances,'' \emph{arXiv preprint arXiv:2211.15876}, 2022.

\bibitem{ramrakhya2023pirlnavpretrainingimitationrl}
R.~Ramrakhya, D.~Batra, E.~Wijmans, and A.~Das, ``Pirlnav: Pretraining with imitation and rl finetuning for objectnav,'' in \emph{CVPR}, 2023.

\bibitem{qin2024langsplat3dlanguagegaussian}
M.~Qin, W.~Li, J.~Zhou, H.~Wang, and H.~Pfister, ``Langsplat: 3d language gaussian splatting,'' in \emph{Proceedings of the IEEE/CVF Conference on Computer Vision and Pattern Recognition}, 2024, pp. 20\,051--20\,060.

\bibitem{qiu2024learninggeneralizablefeaturefields}
\BIBentryALTinterwordspacing
R.-Z. Qiu, Y.~Hu, G.~Yang, Y.~Song, Y.~Fu, J.~Ye, J.~Mu, R.~Yang, N.~Atanasov, S.~Scherer, and X.~Wang, ``Learning generalizable feature fields for mobile manipulation,'' 2024. [Online]. Available: \url{https://arxiv.org/abs/2403.07563}
\BIBentrySTDinterwordspacing

\bibitem{guo2024semanticgaussiansopenvocabularyscene}
\BIBentryALTinterwordspacing
J.~Guo, X.~Ma, Y.~Fan, H.~Liu, and Q.~Li, ``Semantic gaussians: Open-vocabulary scene understanding with 3d gaussian splatting,'' 2024. [Online]. Available: \url{https://arxiv.org/abs/2403.15624}
\BIBentrySTDinterwordspacing

\bibitem{jatavallabhula2023conceptfusionopensetmultimodal3d}
K.~Jatavallabhula, A.~Kuwajerwala, Q.~Gu, M.~Omama, T.~Chen, S.~Li, G.~Iyer, S.~Saryazdi, N.~Keetha, A.~Tewari, J.~Tenenbaum, C.~{de Melo}, M.~Krishna, L.~Paull, F.~Shkurti, and A.~Torralba, ``Conceptfusion: Open-set multimodal 3d mapping,'' \emph{Robotics: Science and Systems (RSS)}, 2023.

\bibitem{gu2023conceptgraphsopenvocabulary3dscene}
Q.~Gu, A.~Kuwajerwala, S.~Morin, K.~M. Jatavallabhula, B.~Sen, A.~Agarwal, C.~Rivera, W.~Paul, K.~Ellis, R.~Chellappa \emph{et~al.}, ``Conceptgraphs: Open-vocabulary 3d scene graphs for perception and planning,'' in \emph{2024 IEEE International Conference on Robotics and Automation (ICRA)}.\hskip 1em plus 0.5em minus 0.4em\relax IEEE, 2024, pp. 5021--5028.

\bibitem{yamazaki2023openfusionrealtimeopenvocabulary3d}
K.~Yamazaki, T.~Hanyu, K.~Vo, T.~Pham, M.~Tran, G.~Doretto, A.~Nguyen, and N.~Le, ``Open-fusion: Real-time open-vocabulary 3d mapping and queryable scene representation,'' in \emph{2024 IEEE International Conference on Robotics and Automation (ICRA)}.\hskip 1em plus 0.5em minus 0.4em\relax IEEE, 2024, pp. 9411--9417.

\bibitem{takmaz2023openmask3dopenvocabulary3dinstance}
A.~Takmaz, E.~Fedele, R.~W. Sumner, M.~Pollefeys, F.~Tombari, and F.~Engelmann, ``{OpenMask3D: Open-Vocabulary 3D Instance Segmentation},'' in \emph{Advances in Neural Information Processing Systems (NeurIPS)}, 2023.

\bibitem{kerbl20233d}
B.~Kerbl, G.~Kopanas, T.~Leimk{\"u}hler, and G.~Drettakis, ``3d gaussian splatting for real-time radiance field rendering.'' \emph{ACM Trans. Graph.}, vol.~42, no.~4, pp. 139--1, 2023.

\bibitem{mildenhall2021nerf}
B.~Mildenhall, P.~P. Srinivasan, M.~Tancik, J.~T. Barron, R.~Ramamoorthi, and R.~Ng, ``Nerf: Representing scenes as neural radiance fields for view synthesis,'' \emph{Communications of the ACM}, vol.~65, no.~1, pp. 99--106, 2021.

\bibitem{kirillov2023segment}
A.~Kirillov, E.~Mintun, N.~Ravi, H.~Mao, C.~Rolland, L.~Gustafson, T.~Xiao, S.~Whitehead, A.~C. Berg, W.-Y. Lo \emph{et~al.}, ``Segment anything,'' in \emph{Proceedings of the IEEE/CVF International Conference on Computer Vision}, 2023, pp. 4015--4026.

\bibitem{zou2023segment}
X.~Zou, J.~Yang, H.~Zhang, F.~Li, L.~Li, J.~Wang, L.~Wang, J.~Gao, and Y.~J. Lee, ``Segment everything everywhere all at once,'' \emph{Advances in Neural Information Processing Systems}, vol.~36, 2024.

\bibitem{xie2024sedsimpleencoderdecoderopenvocabulary}
B.~Xie, J.~Cao, J.~Xie, F.~S. Khan, and Y.~Pang, ``Sed: A simple encoder-decoder for open-vocabulary semantic segmentation,'' in \emph{Proceedings of the IEEE/CVF Conference on Computer Vision and Pattern Recognition}, 2024.

\bibitem{liu2022convnet2020s}
Z.~Liu, H.~Mao, C.-Y. Wu, C.~Feichtenhofer, T.~Darrell, and S.~Xie, ``A convnet for the 2020s,'' in \emph{Proceedings of the IEEE/CVF conference on computer vision and pattern recognition}, 2022, pp. 11\,976--11\,986.

\bibitem{grunnet2019subpixel}
A.~Grunnet-Jepsen, J.~Sweetser, T.~Khuong, D.~Tong, and J.~Woodfill, ``Subpixel linearity improvement for intel{\textregistered}{\textregistered} realsense depth camera d400 series,'' \emph{Retrieved February}, vol.~1, p. 2020, 2019.

\bibitem{hart1968astar}
P.~E. Hart, N.~J. Nilsson, and B.~Raphael, ``A formal basis for the heuristic determination of minimum cost paths,'' \emph{IEEE Transactions on Systems Science and Cybernetics}, vol.~4, no.~2, pp. 100--107, 1968.

\bibitem{wang2023yolov7}
C.-Y. Wang, A.~Bochkovskiy, and H.-Y.~M. Liao, ``{YOLOv7}: Trainable bag-of-freebies sets new state-of-the-art for real-time object detectors,'' in \emph{Proceedings of the IEEE/CVF Conference on Computer Vision and Pattern Recognition (CVPR)}, 2023.

\bibitem{lin2015microsoftcococommonobjects}
T.-Y. Lin, M.~Maire, S.~Belongie, J.~Hays, P.~Perona, D.~Ramanan, P.~Doll{\'a}r, and C.~L. Zitnick, ``Microsoft coco: Common objects in context,'' in \emph{Computer Vision--ECCV 2014: 13th European Conference, Zurich, Switzerland, September 6-12, 2014, Proceedings, Part V 13}.\hskip 1em plus 0.5em minus 0.4em\relax Springer, 2014, pp. 740--755.

\bibitem{Cheng2024YOLOWorld}
T.~Cheng, L.~Song, Y.~Ge, W.~Liu, X.~Wang, and Y.~Shan, ``Yolo-world: Real-time open-vocabulary object detection,'' in \emph{Proc. IEEE Conf. Computer Vision and Pattern Recognition (CVPR)}, 2024.

\bibitem{habitatchallenge2022}
K.~Yadav, S.~K. Ramakrishnan, J.~Turner, A.~Gokaslan, O.~Maksymets, R.~Jain, R.~Ramrakhya, A.~X. Chang, A.~Clegg, M.~Savva, E.~Undersander, D.~S. Chaplot, and D.~Batra, ``Habitat challenge 2022,'' \url{https://aihabitat.org/challenge/2022/}, 2022.

\bibitem{anderson2018evaluationembodiednavigationagents}
\BIBentryALTinterwordspacing
P.~Anderson, A.~Chang, D.~S. Chaplot, A.~Dosovitskiy, S.~Gupta, V.~Koltun, J.~Kosecka, J.~Malik, R.~Mottaghi, M.~Savva, and A.~R. Zamir, ``On evaluation of embodied navigation agents,'' 2018. [Online]. Available: \url{https://arxiv.org/abs/1807.06757}
\BIBentrySTDinterwordspacing

\bibitem{wani2020multion}
S.~Wani, S.~Patel, U.~Jain, A.~X. Chang, and M.~Savva, ``Multion: Benchmarking semantic map memory using multi-object navigation,'' in \emph{NeurIPS}, 2020.

\bibitem{xu2021fastlio2fastdirectlidarinertial}
W.~Xu, Y.~Cai, D.~He, J.~Lin, and F.~Zhang, ``Fast-lio2: Fast direct lidar-inertial odometry,'' \emph{IEEE Transactions on Robotics}, vol.~38, no.~4, pp. 2053--2073, 2022.

\end{thebibliography}
\end{document}